# Cluster-based Method for Eavesdropping Identification and Localization in Optical Links

Haokun Song[1,2]
1 Electrical Engineering Department
Chalmers University of Technology
2 School of Electronic Engineering
Beijing University of Posts and Telecommunications
Gothenburg, Sweden
haokun@ chalmers.se

Rui Lin
Electrical Engineering Department
Chalmers University of Technology
Gothenburg, Sweden
ruilin@chalmers.se

Andrea Sgambelluri
Institute of Telecommunications, Informatics and Photonics
Scuola Superiore Sant'Anna
Pisa, Italy
Andrea.Sgambelluri@santannapisa.it

Filippo Cugini
Institute of Telecommunications, Informatics and Photonics
National, Inter-University Consortium for Telecommunications (CNIT)
Pisa, Italy
filippo.cugini@cnit.it

Yajie Li
School of Electronic Engineering
Beijing University of Posts and Telecommunications
Beijing, China
yajieli@bupt.edu.cn

Jie Zhang
School of Electronic Engineering
Beijing University of Posts and Telecommunications
Beijing, China
jie.zhang@bupt.edu.cn

Paolo Monti
Electrical Engineering Department
Chalmers University of Technology
Gothenburg, Sweden
paolo@chalmers.se

*Abstract*—We propose a cluster-based method to detect and locate eavesdropping events in optical line systems characterized by small power losses. Our findings indicate that detecting such subtle losses from eavesdropping can be accomplished solely through optical performance monitoring (OPM) data collected at the receiver. On the other hand, the localization of such events can be effectively achieved by leveraging in-line OPM data.

*Keywords—eavesdropping, power, machine learning*

## I. INTRODUCTION

Attacks and interference incidents against optical networks are becoming increasingly severe [1]. Physical-layer-based monitoring systems and various detection methods can enhance optical transmission systems' security. Among them, power degradation and the associated degradation in the quality of service is one of the inevitable results of almost all malicious attacks, unexpected failures, and anomalous events. The most commonly used detection methods are based on time domain reflection [2,4] and Rayleigh scattering [3]. Although failure localization has advantages, existing methods are still costly and complex [4]. Monitoring link impairments or channel conditions can also negatively impact signal transmission.

The application of machine learning in optical communications makes it possible to ensure transmission quality while conducting monitoring [5]. There are ample precedents in using machine learning algorithms for the detection of eavesdropping [6] [7], physical layer attacks [8], and fiber perimeter security [9], among others. From a data collection perspective, signal sequences and optical performance monitoring (OPM) parameters can be utilized. Among the various machine learning methods, supervised learning is undoubtedly employed more frequently [6-9]. However, a large amount of labeled data is hard to obtain in practice. On the other hand, unsupervised learning is more suitable for anomaly detection use cases [10] since it eliminates the requirement for prior knowledge of abnormal events. There is research that mixes supervised learning with unsupervised learning for anomaly detection in optical networks [11]. However, in most cases investigated in the literature, the performance parameters vary widely between normal and abnormal events. Taking a step further from anomaly detection, localization of such events is even more significant. Various data sources are utilized to accomplish the localization task. Localization of soft failures of specific devices can be achieved by using the tranceiver parameters and spectrum at multiple nodes [12] [13]. In [14], with the data availability at the receiver end, deep neural-network-based digital backpropagation is utilized to localize abnormal power loss in 15 spans of fibre transmission [14]. Nonetheless, detection and localization of events that have little impact on link performance, e.g., eavesdropping, is still challenging.

There is currently no universally established eavesdropping procedure, making it challenging to determine what should be monitored to detect eavesdropping activities. Nevertheless, a common characteristic among all eavesdropping techniques is the occurrence of power loss. Therefore, the initial necessity to effectively identify eavesdropping attempts is to comprehend various level of power loss. Subsequently, outlining the methodology to distinguish between power loss and instances of eavesdropping becomes imperative.

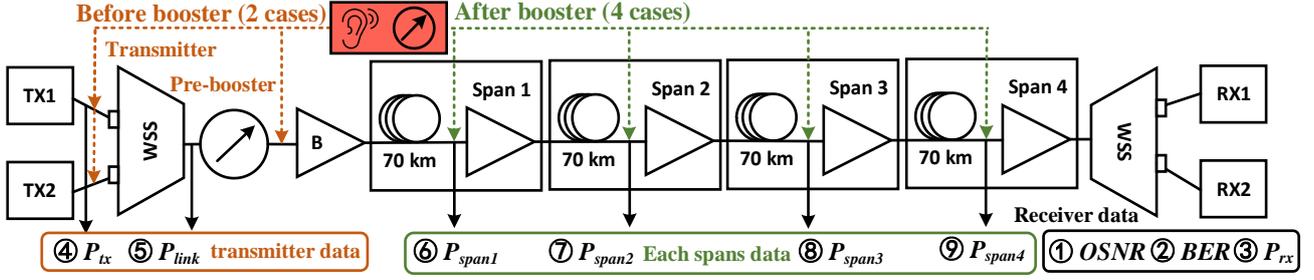

Fig. 1. Flow chart of power-loss eavesdropping simulation system.

In this paper, we present our current focus, lying solely on the initial phase: recognizing and pinpointing instances of power loss through OPM data variation over the whole system with a machine learning method. This study delves into the discernment of varying degrees of eavesdropping severity, ranging from 0.5dB to 3dB power loss. As an illustrative instance of eavesdropping localization, the paper examines explicitly the method of fiber-bending eavesdropping, which entails a power loss of 0.8dB [15]. The data are obtained by a two-channel wavelength division multiplexing (WDM) system simulation, and attenuators emulate the events of different power losses. The data are clustered using an clustering method without training. The detection clustering results demonstrate that the eavesdropping ranging from 0.5dB to 3dB power loss all achieve a 100% label matching rate with receiver OPM data. The sum of squared errors (SSE) at 3dB eavesdropping classification is down to 4.16. Besides, 100% label matching rate localization of fiber-bending eavesdropping can be achieved by leveraging in-line OPM data including of transmitter, receiver, and span data.

## II. POWER-DATALOSS EAVESDROPPING SIMULATION SYSTEM

### A. Simulation setup

The low-loss eavesdropping simulation system in Fig. 2 is built based on the existing experimental platform. All parameters are set to be consistent with the experiment to ensure the credibility of the simulation data. As shown in Fig. 2, the 112Gbps dual-polarization quadrature phase shift keying (DP-QPSK) WDM system with two channels (1550.12 nm and 1550.92 nm) and four fiber spans, which can be divided into four parts: transmitter, link, receiver, and monitor. All parameters at the transmitter are held constant to keep the output power at 0 dBm. The second attenuator in the link is used to simulate different power loss events. The booster gain is adjusted so that the power into the in-line fiber remains at 0 dBm. The attenuator in the loop simulates the additional attenuation of the actual fiber. The in-line amplifiers compensate for all the attenuation in the current span so that the power into the next fiber span can be consistent.

### B. OPM Data Collection

The sample data obtained from each run includes the following 10 optional parameters: optical signal noise ratio (*OSNR*), bit error rate (*BER*), power at the receiver ($P_{rx}$), power at the transmitter ($P_{tx}$), power before the booster ($P_{link}$), and power of each span ($P_{spani, i=1, 2, 3, 4}$). Power can be obtained from the optical power meter. *OSNR* is available from the WDM analyzer. The *BER* is calculated by the *BER* test set located at the transmitter part, which compares the transmitted and received sequences and counts the different bits for *BER* calculation. These parameters can be obtained directly from the monitoring components. In real-world systems, OSNR, BER, and $P_{rx}$ can be acquired directly at the receiver. At the same time, the remaining parameters need to be communicated to the receiver end, e.g., via the control/management plane.

In the simulation, an attenuator serves the purpose of emulating the eavesdropping device. The power loss induced by a typical clip-on coupler-based eavesdropping device [12] is measured as a benchmark to establish the scope of the power range under investigation. Measured by OTDR, a power loss of 0.8dB on the link can be observed, which is shown in Fig. 2. Attenuators are used to emulate eavesdropping of different severity (0, 0.5, 0.8, 1.0, 1.5, 2.0 and 3.0 dB) at various locations (normal operating conditions without eavesdropping, 2 cases before booster and 4 cases after booster, in total 7 cases indicated as dashed arrows in Fig.1). For each case, 200 data samples are collected, where each data sample includes either *OSNR, BER* and $P_{rx}$ only, or all the nine parameters. A total of 1400 data samples are collected from the simulation system, catering for both eavesdropping identification and localization, respectively.

Initially, we consider pre-booster eavesdropping as a case study to investigate the detectability of eavesdropping with different severity levels. This scenario allows the eavesdropper to gain access to the optimal signal quality. Subsequently, fibre-bending eavesdropping that brings 0.8 dB loss is used as an example of eavesdropping localization.

## III. BISECTING K-MEANS CLUSTERING

K-means clustering is a method that aims to partition n observations into k clusters by minimizing within-cluster variances. The bisecting k-means is a type of hierarchical clustering able to overcome the problem K-means has when it converges to a local minimum. The clustering process of bisecting k-means is shown in Fig. 3:

- The standardized data and the number of clustering categories K are provided.

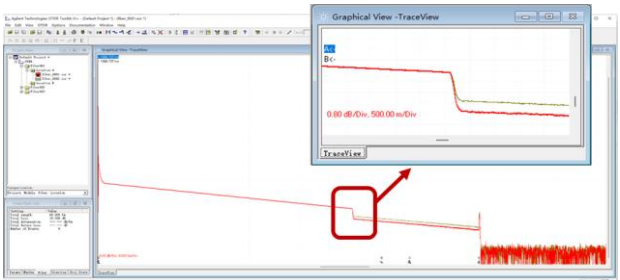

Fig. 2. The optical power loss due to the clip-on coupler.

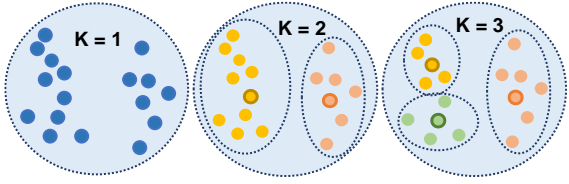

Fig. 3. Schematic diagram of bisecting k-means clustering.

- Treat all the data points as a cluster and make the points clustered into two categories.
- Select the cluster with the largest inertia (i.e., the largest SSE) and continue to divide it in two.
- Stop when the number of clusters equals K.

The data collected in this scheme includes features and labels, which can be classified with supervised learning in theory. There are several reasons for choosing bisecting k-means. Firstly, a large amount of data with definite labels required for supervised learning is challenging to obtain in real-world systems, while data labelling can be mitigated in unsupervised learning. Secondly, this algorithm does not require prior training of the model, and the classification results can be obtained directly. Finally, the selected bisecting k-means is more efficient than plain k-means and excludes the uncertainty of random initialization. Besides, the true labels in the data can help evaluate the clustering algorithm's accuracy. It makes up for the disadvantage that unsupervised learning results are difficult to assess. The performance indicators involved in this work are label matching rate and SSE:

$$\text{SSE} = \sum_{i=1}^{n} \sum_{j=1}^{K} w^{(i,j)} = \|x^i - center^j\|_2^2 \quad (1)$$

where $j$ represents the cluster number and the total $n$ points within the cluster are denoted by $i$. w is the Euclidean distance. $center^j$ indicates the center of cluster $j$.

## IV. RESULTS AND DISCUSSION

### A. Identification of eavesdropping with different severity levels

The extent of power loss caused by eavesdropping varies, as different methods and operations can yield varying levels of severity. Consequently, the resultant power losses also differ significantly. Nonetheless, it is confirmed that the detection becomes more challenging as the power loss decreases. Thus, the primary objective is to comprehend the threshold of detectable power loss.

Firstly, data samples containing OSNR, BER, and $P_{rx}$ are used only to detect eavesdropping. The transmission parameters under normal operating conditions without eavesdropping, i.e., without extra loss due to eavesdropping, are used as the benchmark, and each event is clustered with $k=2$, respectively. The identification of eavesdropping can be made with only the three receiver data. The result is shown in Figure 4. The higher the power loss, the lower the SSE after completing the classification. It indicates that the classification results are explicit and effective. The classification results in both cases can achieve a 100% label matching rate. The limited fluctuation in the simulation system leads to a concentration of OPM data for different power loss events. Therefore, we expect better classification results in the simulation than in an experimental system.

### B. Localization of eavesdropping with different OPM data

A total of 6 eavesdropping locations were set up, including the transmitter, pre-booster, and 4 spans. Together with the case of no eavesdropping, there were 7 other cases. As shown in Fig. 2, the seven scenarios can be grouped into three main clusters: normal conditions without eavesdropping, before, and after booster eavesdropping. As shown in Fig. 5, if only the OPM parameters of the receiving end are available, all data can be roughly classified (three main clusters). If the transmitter end parameters are obtained, a further distinction can be made between the two cases of transmitter end and pre-booster. Similarly, if the in-line parameters can be obtained, it is possible to distinguish further in which span the eavesdropping occurs. In summary, the more in-line equipment provides data, the more precise the localization results available.

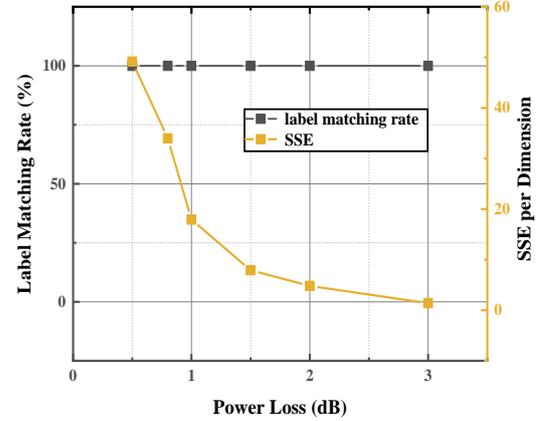

Fig. 4. Diagram of the classification performance in different power loss eavesdropping.

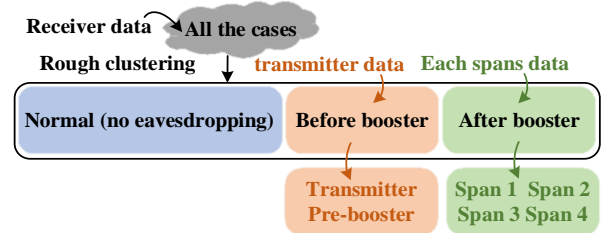

Fig. 5. Diagram of the usage of different OPM data.

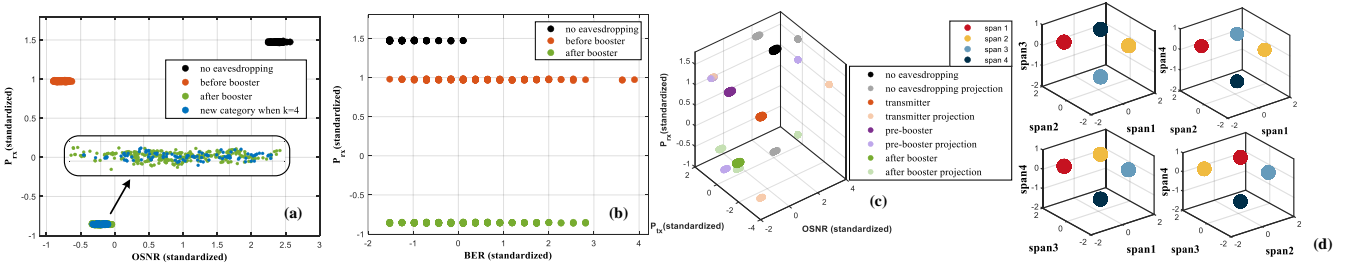

Fig. 6. Diagram of the location performance with different parameters (a) rough clustering with $OSNR$ and $P_{rx}$ (b) rough clustering with $BER$ and $P_{rx}$ (c) before booster clustering with $OSNR$, $P_{tx}$ and $P_{rx}$ (d) before booster clustering with 4 kind of spans data combinations.

TABLE I. CLUSTERING RESULTS FOR DIFFERENT PARAMETERS

| OPM data | | | Clustering Results | |
|---|---|---|---|---|
| | | k | Label Matching Rate | SSE/Dimension |
| Rough clustering | ①②③ | 3 | 100% | 529.54/3≈176.51 |
| | ①③ | 3 | 100% | 23.47/2≈11.74 |
| Before booster | ①②③④ | 4 | 100% | 529.44/4=132.36 |
| | ①②③⑤ | 4 | 58.43% | 330.89/4≈82.72 |
| | ①③④ | 4 | 100% | 23.59/3≈7.86 |
| After booster | ⑥⑦⑧⑨ | 4 | 100% | 0.17/4≈0.04 |
| | ⑥⑦⑧ | 4 | 100% | 5.20/2=2.60 |

The localization results are shown in Table I. The algorithm performs a K=3 rough clustering when only the OPM parameters of the receiving end are available. The result is given in Fig. 6 (a) when the algorithm is made to categorize further, *i.e.*, K=4. The 4 cases of data after booster (4 spans) are divided randomly into two categories, indicating that the clustering results of this step are no longer meaningful. In addition, the clustering results with the parameter BER are given in Fig. 6 (b). Compared with the classification results with the parameter *OSNR* in (a), the *BER* is more dispersed and is less beneficial for clustering. After streamlining the parameters by only utilizing the *OSNR* and the $P_{rx}$ for classification, the SSE decreases significantly, which is the best result for rough clustering. It is further speculated that the cause of the *BER* results may be that the *BER* of the simulation software is not estimated from the channel conditions but is calculated from the number of statistical BER bits. Therefore, every bit error significantly affects the *BER* calculation at this BER level. The BER in an actual deployment may be a more reliable parameter.

Including the transmitter power parameter allows the distinction between transmitter eavesdropping and pre-booster eavesdropping. And it is the only parameter that successfully distinguishes between the two. The results are shown in Fig. 6 (c). The SSE/dim after streamlining the parameters drops to 7.86. Based on the rough classification, the parameters of different spans in the link can help further locating the eavesdropper. The power of each span reflects the presence or absence of eavesdropping within this span, so at least three parameters are needed to distinguish the four cases fully. The results for any three parameters combination are shown in Fig. 6 (d). Presumably, at least N-1 parameters are needed for N-span segments.

## V. CONCLUSION

A cluster-based method for eavesdropping detection and localization is proposed and validated. It requires only OPM data and achieves a 100% label matching rate for 0.5-3 dB loss eavesdropping. In the localization analysis with 0.8dB eavesdropping, 100% label matching rate localization can be achieved by possessing the transmitter end, receiver end, and spans data. Validation of this method in an experimental system and attempting more challenging fuzzy clustering are envisioned for future work.

ACKNOWLEDGMENT

This work is supported by the EUREKA cluster CELTIC-NEXT project AI-NET PROTECT funded by VINNOVA, the Swedish Innovation Agency.